\documentclass[11pt,a4paper]{article}
\usepackage{authblk}
\usepackage[hyperref]{emnlp2020}
\usepackage{times}
\usepackage{latexsym}

\usepackage{url}
\usepackage{graphicx}
\usepackage{amsmath}
\usepackage{booktabs}
\usepackage{algorithm}
\usepackage{algorithmic}
\usepackage{mathrsfs}
\usepackage{amssymb}
\usepackage{setspace}
\usepackage{comment}
\usepackage{multirow}
\usepackage{bbm}
\newtheorem{defn}{Definition}[section]

\newlength\myindent
\newcommand\bindent{%
  \begingroup
  \setlength{\itemindent}{\myindent}
  \addtolength{\algorithmicindent}{\myindent}
}
\newcommand\eindent{\endgroup}

\usepackage{microtype}

\aclfinalcopy 

\title{FedE: Embedding Knowledge Graphs in Federated Setting}

\author[1]{\textbf{Mingyang Chen}}
\author[1]{\textbf{Wen Zhang}}
\author[2]{\textbf{Zonggang Yuan}}
\author[3]{\textbf{Yantao Jia}}
\author[1 *]{\textbf{Huajun Chen}}
\affil[1]{College of Computer Science and Technology, Zhejiang University}
\affil[2]{Huawei NAIE CTO Office}
\affil[3]{Huawei Technologies Co., Ltd}
\affil[ ]{\{mingyangchen, wenzhang2015, huajunsir\}@zju.edu.cn}
\affil[ ]{yuanzonggang@huawei.com}
\affil[ ]{jamaths.h@163.com}

\newcommand\blfootnote[1]{%
\begingroup 
\renewcommand\thefootnote{}\footnote{#1}%
\addtocounter{footnote}{-1}%
\endgroup 
}

\date{}

\begin{document}
\maketitle
\begin{abstract}

Knowledge graphs (KGs) consisting of triples are always incomplete, so it's important to do Knowledge Graph Completion (KGC) by predicting missing triples.
Multi-Source KG is a common situation in real KG applications which can be viewed as a set of related individual KGs where different KGs contains relations of different aspects of entities.
It's intuitive that, for each individual KG, its completion could be greatly contributed by the triples defined and labeled in other ones.
However, because of the data privacy and sensitivity, a set of relevant knowledge graphs cannot complement each other's KGC by just collecting data from different knowledge graphs together. 
Therefore, in this paper, we introduce federated setting to keep their privacy without triple transferring between KGs and apply it in embedding knowledge graph, a typical method which have proven effective for KGC in the past decade.
We propose a Federated Knowledge Graph Embedding framework FedE, focusing on learning knowledge graph embeddings by aggregating locally-computed updates. 
Finally, we conduct extensive experiments on datasets derived from KGE benchmark datasets and results show the effectiveness of our proposed FedE.
\blfootnote{* \, Corresponding author.}


\end{abstract}

\section{Introduction}


Knowledge Graph (KG) represents the fact that entity \textit{h} and \textit{t} are connected by relation \textit{r}
with the form of triple \textit{(head entity, relation, tail entity)}, \textit{(h, r, t)} in short.
Currently, a plenty of large-scale knowledge graphs are proposed
, such as YAGO \cite{yago}, Freebase \cite{freebase}, NELL \cite{nell} and Wikidata \cite{wikidata}, and a lot of downstream tasks can take advantage of knowledge graph. However, many knowledge graphs are incomplete, so it's of great significance to predict missing triples based on current triples, namely Knowledge Graph Completion (KGC).

\begin{figure}[t]
\centering
\includegraphics[scale=0.34]{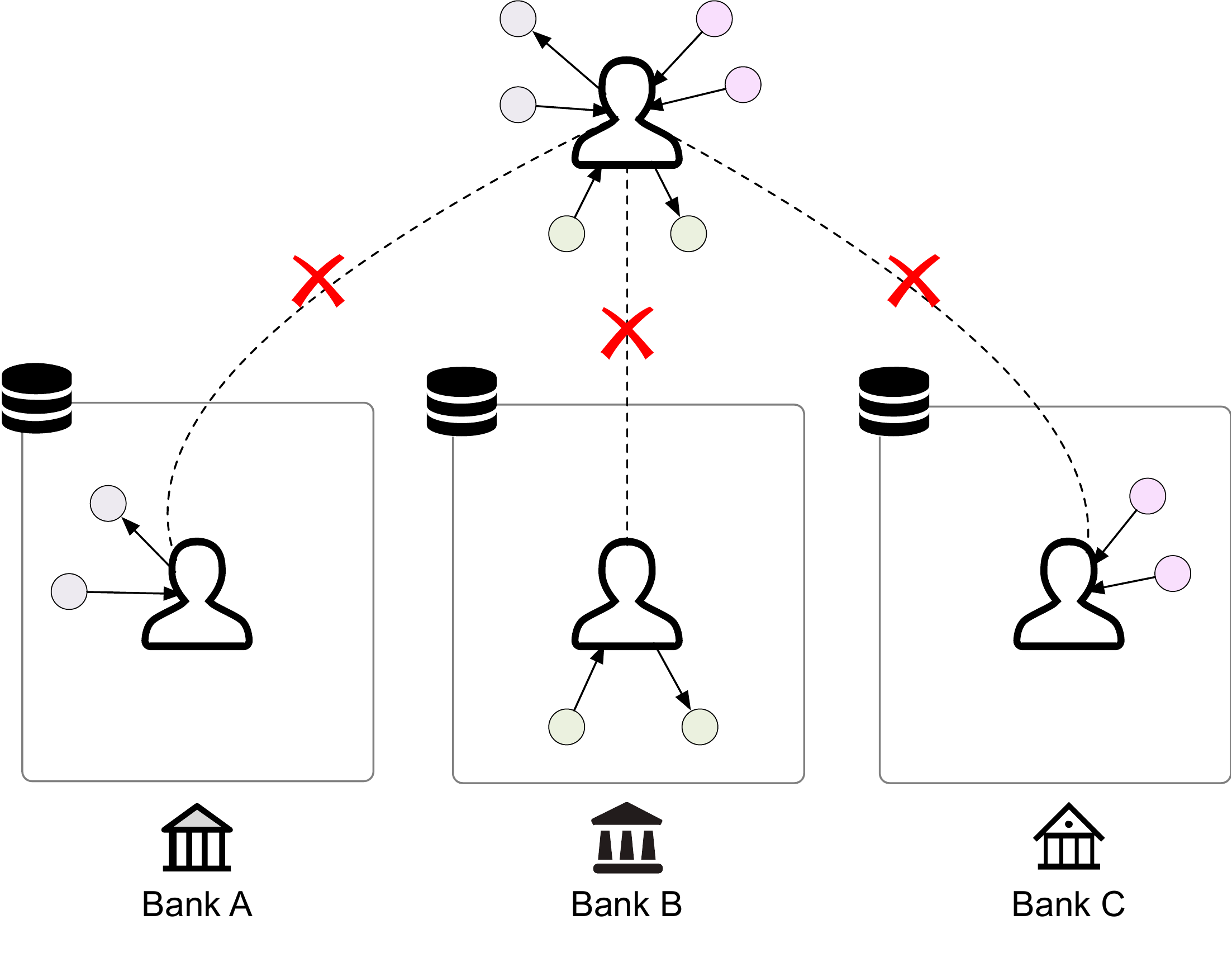}
\caption{Challenge of Multi-Source KG. A person (entity) can be involved in different banks and generates related data (triples), while each bank is not willing to expose their data to get better entity representation for further applications.}
\label{fig:intro}
\end{figure}

In real KG applications, it's common that entities are involved in different related KGs, and we call such situation as Multi-Source KG. Knowledge graph completion for each individual KG in such situation is also important. For example, as shown in Figure \ref{fig:intro}, a person (entity) has different operations in different banks, and each bank will save related data (triples) for this person. It's intuitive that, for KG in each bank, relations defined and labeled (i.e. triples) in other KGs have potential of improving the quality of representation for this person (entity) and enhancing the KGC performance. However, KGs often cannot complement each other's KGC capabilities by just collecting triples from different KGs together, because of the data privacy and sensitivity, especially in some specific application domains (e.g. financial and medical domain) and under some regulations (e.g. European Union General Data Protection Regulation). How to take advantage of the complementary capabilities of different related knowledge graphs while keeping data privacy is an urgent problem in real KG applications.


In order to solve this problem, we resort to Federated Learning \cite{FedAvg} which is a decentralized method proposed to protect privacy by training a shared model without collecting data from different clients/devices/platforms. 
The model training is conducted by aggregating locally-computed updates and the data in clients won't be transferred to anywhere for data privacy. Nowadays, Knowledge Graph Embedding (KGE) methods which focus on learning reasonable embeddings (vector representations) for relations and entities have proven to be effective to do knowledge graph completion, and lots of embedding-based methods \cite{TransE, TransH, TransR, DistMult, ComplEx, CrossE, IterE, RotatE} for KGC have been proposed in the past decade.

To take advantage of different relevant knowledge graphs and ensure the data privacy, inspired by the conception of federated learning and knowledge graph embedding, we embed knowledge graphs in federated setting in this work. Specifically, we propose a Federated Knowledge Graph Embedding framework, \textbf{FedE}, to leverage the capability of knowledge graph embeddings for predicting missing triples by learning shared entity embeddings without collecting triples from different clients to a central server. There are two kinds of roles in our FedE, the server and client. 
Each client has its own knowledge graph and different clients have overlapping entity sets but not aware of others' relation sets and triples. The main idea of our FedE is that training embeddings only on each client and aggregating entity embeddings on server while not collecting triples of clients to the central server for privacy protection. For each client, triples and relation set won't be exposed to others, and entity set will only be read by the central server, so triples cannot be inferred by other clients and the data privacy can be guaranteed. 
Furthermore, for fusion the capability of learned embeddings based only on one client and in federated setting, we also design a model fusion procedure on each client.

We conduct experiments on four knowledge graphs derived from two benchmark knowledge graph datasets, and show the effectiveness of our FedE in federated knowledge graph setting which not only achieves significant improvement compared to conducting KGC within single KG but also achieves results comparable with or even better than results from the setting of collecting entire data together.
In summary, main contributions of our work are as follows:
\begin{itemize}
\item We consider the federated setting in the knowledge graph and formulate its completion task as a new task, i.e. federated knowledge graph completion;
\item We propose a general framework FedE to embed knowledge graphs in federated setting, which could be applied to any methods containing entity embeddings;
\item We experimentally prove that our proposed FedE is effective on four datasets derived from conventional KGE benchmarks.
\end{itemize}

\section{Federated Knowledge Graph Embedding}

\subsection{Task Formulation}
We formally define the Knowledge Graph at the beginning, and then, we extend such definition to Federated Knowledge Graph Completion, the task which we are interested in.
\begin{defn}
	(Knowledge Graph $\mathcal{G}$) A knowledge graph is defined as $\mathcal{G} = \{ \mathcal{E}, \mathcal{R},  \mathcal{T}\}$. $\mathcal{E} = \{e_i\}_{i=1}^{n}$ is a set which consists of $n$ entities and $\mathcal{R} = \{r_i\}_{i=1}^m$ is the a set which consists of $m$ relations. A triples set $\mathcal{T} = \{ (h, r, t)\in \mathcal{E} \times \mathcal{R} \times \mathcal{E}\}$ models the relation between different entities. 
\end{defn}


\begin{defn}
\label{def:fedkg}
    (Federated Knowledge Graph Completion)
	This task focuses on doing knowledge graph completion in federated setting. Specifically, in such setting, $C$ knowledge graphs $\{\mathcal{G}_{c}\}_{c=1}^{C} = \{\{ \mathcal{E}_c, \mathcal{R}_c,  \mathcal{T}_c\}\}_{c=1}^{C}$ locate in different clients with overlapping entity sets, and each knowledge graph defines its own relation set and triples.

	This completion task focuses on predicting missing triples like $(h, r, ?)$ in each KG by taking advantage of information from other KGs in federated setting. For each KG in such setting, the triples and relation set are not exposed to others, and the entity sets of different clients can only be read by a central server which will maintain a overall entity table to record all the unique entities from different clients and can map entities from different clients to this table. The unawareness of relations, entities and triples in each other client guarantee the data privacy for each client.
	
\end{defn}

\subsection{FedE}

In this section, we propose a Federated Knowledge Graph Embedding framework FedE to solve the Federated Knowledge Graph Completion task as mentioned above. Following previous FederatedAveraging (FedAvg) method \cite{FedAvg}, we model our FedE as two parts, a central server and a set of clients. Under the definition above, different knowledge graphs locate in different clients with overlapping entity sets, and define there own triples and relation sets. Knowledge graphs in different clients using there own triples to update the entity and relation embeddings. A central server controls the aggregation and sharing of global entity embeddings. This server is responsible of maintain a overall entity table to record all the unique entities from different clients and can map entities from different clients to this table.

\begin{figure*}[t]
\centering
\includegraphics[scale=0.48]{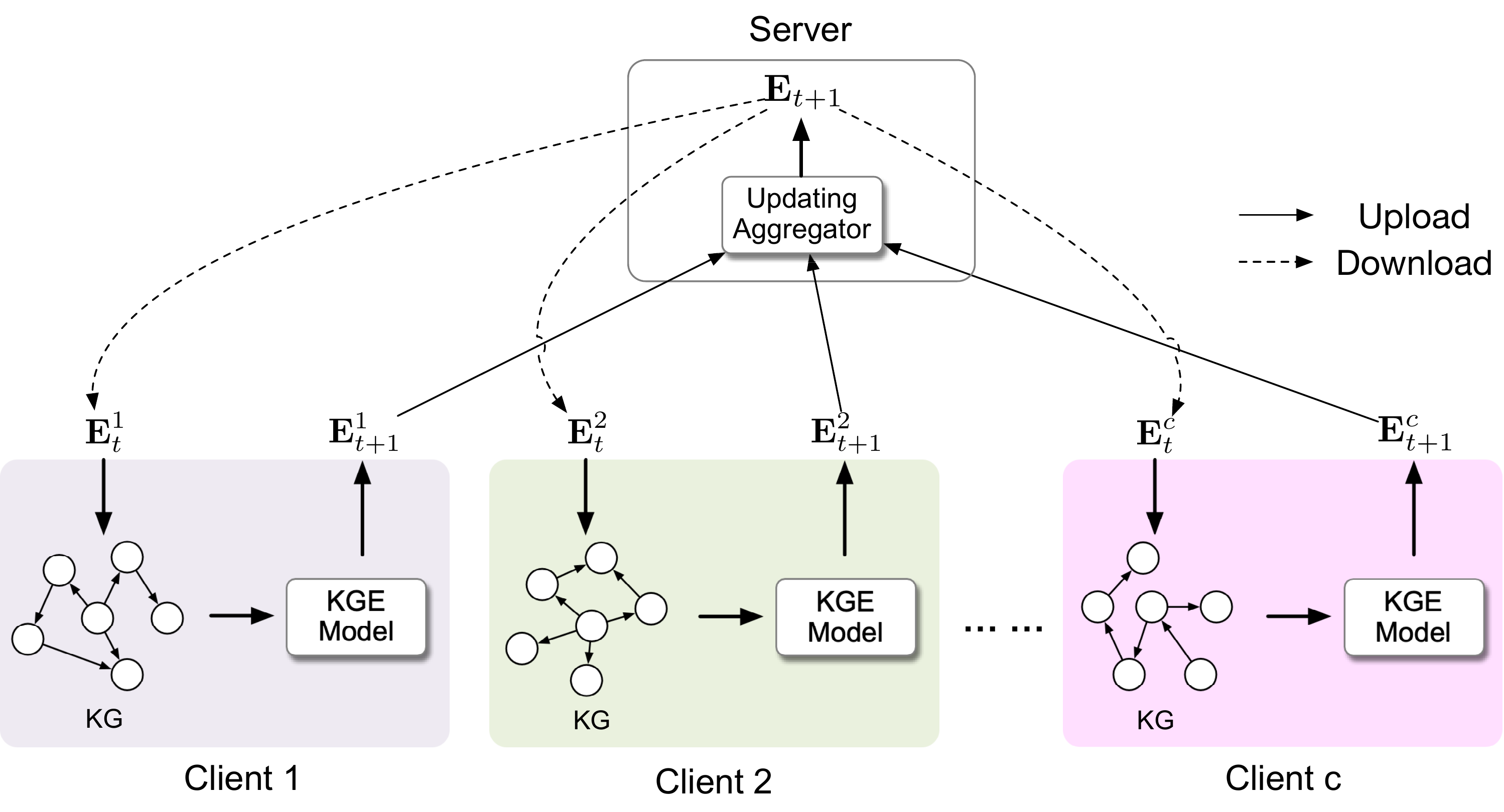}
\caption{Overview of FedE. Illustrating the procedure of client updating and server operation at round $t$.}
\label{fig:overview}
\end{figure*}

We firstly introduce the procedure of sever operation and client updating, and the final model fusion procedure tries to fusion the embeddings learned with and without federated setting to lift the capability of embeddings from different aspects. The overview of FedE is shown in Figure \ref{fig:overview}. The overall procedure of our framework is described in Algorithm \ref{alg:FedE}.

\begin{algorithm}[tb]
\setstretch{1}
\caption{FedE Framework}
\label{alg:FedE}
\begin{algorithmic}
\REQUIRE The number of clients $C$; The faction of clients selected in each round $F$; The number of local epoch $E$ and local batch size $B$; The learning rate $\eta$.
 
\vspace*{1\baselineskip} 
\STATE {\bf Server:}
\bindent
\STATE Server constructs permutation matrices $\{\mathbf{P}^c\}_{c=1}^{C}$ and existence vectors $\{\mathbf{v}^c\}_{c=1}^{C}$ and initializes $\mathbf{E}_0$.
\FOR{round $t=0,1,\dots$}
    \STATE Server distributes $\mathbf{E}_t$ to each client.
	\STATE $\mathcal{C}_t$ $\leftarrow$ (random set of $F \times C$ clients)
	\FOR{each client $c \in \mathcal{C}_t$ {\bf in parallel}}
		\STATE $\mathbf{E}^c_{t+1} \leftarrow$ {\bf ClientUpdate}$(c, \mathbf{P}^{c\top} \mathbf{E}_t)$
	\ENDFOR
	\STATE $\mathbf{E}_{t+1} \leftarrow \left( \mathbbm{1} \oslash \sum \mathbf{v}^{c} \right) \otimes \sum \mathbf{P}^{c} \mathbf{E}^{c}_{t+1}$
\ENDFOR

\FOR{each client $c$ from 1 to $C$ {\bf in parallel}}
\STATE Do model fusion on client $c$
\ENDFOR

\eindent

\vspace*{1\baselineskip}
\STATE {\bf ClientUpdate}{($c, \mathbf{E}$)}:
\bindent
\FOR{each local epoch $e$ from 1 to $E$}
	\STATE $\mathcal{B} \leftarrow$ (random split triples $\mathcal{T}_c$ into batches of size $B$)
	\FOR{batch $b \in \mathcal{B}$}
		\STATE Compute loss $\mathcal{L}$ for $b$ by Eq. \ref{eq:loss}.
		\STATE $\mathbf{E} \leftarrow \mathbf{E} - \eta \nabla \mathcal{L} $
	\ENDFOR
\ENDFOR
\STATE return $\mathbf{E}$
\eindent

\end{algorithmic}
\end{algorithm}

\subsubsection{Server Operation}

The server in our FedE framework is responsible of aggregating entity embeddings from different clients and do average operation, finally, send aggregated entity embeddings back to each clients. We call one time of aggregation as one round for FedE.

Because of the overlapping entity sets of different clients, at the very beginning, entity sets from different clients will be sent to the central server, and the server will record all the unique entities and maintain a entity table itself. By the way, a set of permutation matrices $\{\mathbf{P}^c\ \in \{0,1\}^{n \times n_c}\}_{c=1}^{C}$ and existence vectors $\{\mathbf{v}^c \in \{0,1\}^{n \times 1} \}_{c=1}^{C}$ will be constructed for aggregating entity embeddings from clients, where $C$ is the total number of clients, $n$ is the number of all unique entities and $n_c$ is the number of entities from client $c$. Specifically, the permutation matrix $\mathbf{P}^c$ is used for mapping entity matrix from client $c$ to the server's entity table, where $\mathbf{P}^c_{i,j} = 1$ if the $i$-th entity in entity table corresponds to the $j$-th entity from client $c$. $\mathbf{v}^{c}_i = 1$ indicates that the $i$-th entity in entity table exists in client $c$.

Next, server will randomly initialize entity embedding matrix $\mathbf{E}_0 \in \mathbb{R}^{n \times d_e}$ based on above entity table, where $d_e$ is the dimension of entity embeddings. A permutation will be conducted before sending entity embeddings to client each client, 
\begin{equation}
\begin{aligned}
	\mathbf{E}_{0}^{c} \leftarrow \mathbf{P}^{c\top} \mathbf{E}_{0} .
\label{eq:permutation}
\end{aligned}
\end{equation}

At round $t$, server firstly send entity embedding matrix $\mathbf{P}^{c\top} \mathbf{E}_t \in \mathbb{R}^{n_c \times d_e}$ to each client. And $F \times C$ clients will be chosen randomly, where $F$ is the fraction of clients selected in each round, i.e. client fraction. After training embeddings with specific number of epoch in each selected clients, server aggregates entity embeddings $\mathbf{E}_{t+1}^c$ from client $c$ as follows:
\begin{equation}
\begin{aligned}
	\mathbf{E}_{t+1} \leftarrow \left( \mathbbm{1} \oslash \sum_{c=1}^{C} \mathbf{v}^{c} \right) \otimes \sum_{c=1}^{C} \mathbf{P}^{c} \mathbf{E}^{c}_{t+1} ,
\label{eq:ent-avg}
\end{aligned}
\end{equation}

where $\mathbbm{1}$ denotes all-one vector, $\oslash$ denotes element-wise division for vectors and $\otimes$ denotes element-wise multiply with broadcasting, i.e. $[\mathbf{v} \otimes \mathbf{M}]_{i,j} = \mathbf{v}_{i} \times \mathbf{M}_{i,j}$. The Equation \ref{eq:ent-avg} demonstrates the procedure of entity embedding aggregating, where entity embedding matrices from different clients will be permuted by corresponding permutation matrices and the average will be weighted the number of existence of each entity calculated by existence vector.

\subsubsection{Client Updating}

\begin{table}[t]
\linespread{1.4}
\small
\centering

\begin{tabular}{ccc}
\toprule

Model & Score Function & Vector Space \\

\midrule

TransE & $-|| \mathbf{h} + \mathbf{r} - \mathbf{t} ||$ & $ \mathbf{h}, \mathbf{r}, \mathbf{t} \in \mathbb{R}^d $ \\
DistMult & $\mathbf{h}^\top \operatorname{diag}(\mathbf{r}) \mathbf{t}$ & $ \mathbf{h}, \mathbf{r}, \mathbf{t} \in \mathbb{R}^d $ \\
ComplEx & $\operatorname{Re}(\mathbf{h}^\top \operatorname{diag}(\mathbf{r}) \overline{\mathbf{t}})$ & $ \mathbf{h}, \mathbf{r}, \mathbf{t} \in \mathbb{C}^d $ \\
RotatE & $-|| \mathbf{h} \circ \mathbf{r} - \mathbf{t} ||$ & $ \mathbf{h}, \mathbf{r}, \mathbf{t} \in \mathbb{C}^d $ \\

\bottomrule
\end{tabular}

\caption{Score function $f_r(h, t)$ of typical knowledge graph models. $\mathbf{h}, \mathbf{r}, \mathbf{t}$ are embeddings correspond to $h, r, t$. $\operatorname{Re(\cdot)}$ denotes the real vector component of a complex valued vector. $\circ$ denotes the Hadamard product.}
\label{tab:score-function}
\end{table}
 
Following knowledge graph embedding methods, after receiving entity embedding $\mathbf{E}_t^c$ from server at round $t$ for each client $c$, a selected set of clients use there own triples and relation embeddings to update entity embeddings, at the same time, relation embeddings in each client will also be updated.

For a triple $(h, r, t)$ in knowledge graph $\mathcal{G}_c$ in client $c$, we calculate the truth value by score function $f_r(h, t)$, and this score function can be changed by the choice of knowledge graph embedding method. Score functions of four typical KGE methods are described in Table \ref{tab:score-function}. Following the loss function proposed by \citet{RotatE}, the loss for triple $(h, r, t)$ in client $c$ is,
\begin{equation}
\begin{aligned}
	L(h,r,t) =  -\log \sigma (f_r(h, t) - \gamma ) \\  
		 -\sum_{i=1}^{n} p(h, r, t^\prime_i) \log \sigma(\gamma - f_r(h, t^\prime_i)) ,
\label{eq:loss}
\end{aligned}
\end{equation}
where $\gamma$ is the margin which is a hyperparameter. $(h, r, t^\prime_i)$ is a negative sample corresponding to $(h,r,t)$, where $(h, r, t^\prime_i) \notin \mathcal{G}_c$, and $n$ is the number of negative samples. $p(h, r, t^\prime_i)$ is the weight for corresponding negative sample, and this non-uniform negative sampling is called self-adversarial negative sampling, the weight is defined as follows:
\begin{equation}
\begin{aligned}
	p(h, r, t^\prime_j) = \frac{\exp(\alpha f_r(h, t^\prime_j))}{\sum_{i} \exp (\alpha f_r(h, t^\prime_i))} ,
\label{eq:neg-sample}
\end{aligned}
\end{equation}
where $\alpha$ is the temperature.

After $P$ epoches of training on client $c$ at the round $t$, updated entity embeddings $\mathbf{E}^c_{t+1}$ will be uploaded to the sever. 

\subsubsection{Model Fusion}
\label{sec:model-fusion}

It's intuitive that the embeddings learned by federated knowledge graph embedding framework are complementary to the trained embeddings based only on one knowledge graph without federated setting. For this reason, we design a model fusion procedure for knowledge graph embeddings on each client to fusion the embeddings learned with and without federated setting.

To fusion two models, for a triple $(h, r, t)$, we firstly concat score $f_r^s(h, t)$ and $f_r^f(h, t)$ as a feature vector, where score function superscripted by $s$ using embeddings trained by single knowledge graph while score function superscripted by $f$ using embeddings trained in federated setting. Then, using a linear classifier taking such feature vector as input and output the final score for this triple,

\begin{equation}
\begin{aligned}
	s_{(h,r,t)} &= \mathbf{W}\mathbf{x}+\mathbf{b}, \\
	\mathbf{x} &= [f_r^s(h, t); f_r^f(h, t)].
\label{eq:fusion-linear}
\end{aligned}
\end{equation}
This linear classifier is trained by a margin ranking loss to make the positive triples be ranked higher than the negative triples, and the model fusion loss for $(h, r, t)$ in $\mathcal{G}_c$ is defined as follows:

\begin{equation}
\begin{aligned}
	L_f(h,r,t) = \max(0, \beta - s_{(h,r,t)} + s_{(h,r,t^\prime)})
\label{eq:fusion-loss}
\end{aligned}
\end{equation}
where $\beta$ is the margin and $(h,r,t^\prime)$ is a the negative triple. The training object of model fusion procedure is to minimize the model fusion loss of all triples.

\section{Experiments}

In this section, we have following questions to be explored. 1) Can our FedE get better performance than KGE-based methods without federated setting? 2) What are the comparison results between FedE and KGE-based methods with collecting triples from different clients which violate the data privacy? 3) Is FedE effective with the increase of the number of clients? 4) Is the model fusion procedure effective? 5) The impact of different factors (i.e. multi-client parallelism and client computation) on training and testing for FedE.\footnote{The source code could be found at \url{https://github.com/AnselCmy/FedE}.}

\subsection{Datasets and Evaluation Metrics}

To test the effectiveness of our FedE model, we use two benchmark knowledge graph datasets, FB15k-237 \cite{fb15k237} and NELL-995 \cite{DeepPath}. These two benchmark datasets are originally proposed to evaluate knowledge graph embedding methods, in order to evaluate our FedE in federated setting, we create several new datasets in federated setting by distributing triples into clients. Specifically, according to Definition \ref{def:fedkg}, we randomly select relations into clients, and distribute triples into clients according to selected relations. We randomly split FB15k-237 into 3,5,10 clients as FB15k-237-Fed3, -Fed5, -Fed10 and NELL-995 into 3 clients as NELL-995-Fed3, the statistics of these datasets are given in Table \ref{tab:dataset-statistic}. The proportion for training/validation/testing triples is 0.8/0.1/0.1.  

\begin{table}[t]
\linespread{1.4}
\small
\centering

\begin{tabular}{ccccc}
\toprule

Dataset & \#C & \#Rel & \#Ent & \#Tri \\

\midrule

FB15k-237-Fed3 & 3 & 79 & 12595.3 & 103373 \\
FB15k-237-Fed5 & 5 & 47.4 & 11260.4 & 62023.2 \\
FB15k-237-Fed10 & 10 & 23.7 & 8340.1 & 31011.6 \\
NELL-995-Fed3 & 3 & 66.7 & 33453.7 & 51404.3 \\

\bottomrule
\end{tabular}

\caption{Statistics of datasets. \#C denotes the number of clients. \#Rel, \#Ent and \#Tri denote the average of the number of unique relations, entities and triples in clients.}
\label{tab:dataset-statistic}
\end{table}

Following conventional evaluation metrics on knowledge graph embedding methods, we report Mean Reciprocal Rank (MRR) and Hits at N (Hits@N) to evaluate the performance of link prediction on each client. The link prediction ranking results are evaluated in the filtered setting \cite{TransE}, where we filter the candidates triples that appear either in training, validation and testing set when we do evaluation.

\subsection{Implementation}

To demonstrate the capability of our FedE framework, we contrast FedE with knowledge graph embedding method under the sitting \textit{Single} and \textit{Entire}. For setting \textit{Single}, the knowledge graph embeddings are trained only on its own knowledge graph for each client. For setting \textit{Entire}, we collect triples of knowledge graphs from different clients together, and train knowledge graph embeddings on such entire triples, note that this setting violates the privacy protection because triples from different clients are exposed during the procedure of collection. For setting \textit{FedE}, we train knowledge graph embeddings using our proposed FedE in the federated setting and triples on each clients won't be transferred to anywhere. We choose four typical knowledge graph embedding methods, TransE\cite{TransE}, DistMult\cite{DistMult}, ComplEx\cite{ComplEx} and RotatE\cite{RotatE} for setting \textit{Single}, \textit{Entire}, \textit{FedE}.

For setting \textit{FedE}, unless otherwise noted, the number of local epoch $E$ is 3 and the local batch size $B$ is 512. The number of clients chosen in each round, i.e. $k$, for FB15k-237-Fed3 and NELL-995-Fed3 is 3, for FB15k-237-Fed5 and -Fed10 is 5. Adam \cite{adam} with initial learning rate as 0.001 is used for stochastic optimization of client updating. The embedding dimension is 256 for relations and entities. We set margin $\gamma$ and $\beta$ as 10, temperature for self-adversarial negative sampling $\alpha$ as 1 and the number of negative sampling as 256. During training, model will be evaluated on validation set each 10 epochs for setting \textit{Single} and \textit{Entire}, and each 5 rounds for setting \textit{FedE}. We use early stopping with 15 patient epochs/rounds, where the training procedure will be aborted after consecutive 15 epochs/rounds dropping on MRR of validation.

\subsection{Main Results}

\begin{table*}[t]
\linespread{1.3}
\centering
\resizebox{\textwidth}{31mm}{
\begin{tabular}{ccllll|llll|llll}
\toprule

\multirow{2}{*}{KGE} & 
\multirow{2}{*}{Setting} & 
\multicolumn{4}{c}{Client 1} & \multicolumn{4}{c}{Client 2} & \multicolumn{4}{c}{Client 3} \\ 
\cmidrule{3-14}
& & \multicolumn{1}{c}{MRR} & \multicolumn{1}{c}{H@1} & \multicolumn{1}{c}{H@5} & \multicolumn{1}{c}{H@10} & \multicolumn{1}{c}{MRR} & \multicolumn{1}{c}{H@1} & \multicolumn{1}{c}{H@5} & \multicolumn{1}{c}{H@10} & \multicolumn{1}{c}{MRR} & \multicolumn{1}{c}{H@1} & \multicolumn{1}{c}{H@5} &  H@10 \\ 

\midrule

\multirow{3}{*}{TransE} & Single & 
0.5035 & 0.3975 & 0.6362 & 0.6982 & 0.3869 & 0.2561 & 0.5464 & 0.6373 & 0.3372 & 0.2056 & 0.4887 & 0.5967 \\ 
& Entire & 
0.5424 & 0.4345 & 0.6739 & 0.7414 & 0.4098 & 0.2752 & 0.5756 & 0.6689 & 0.3568 & 0.2202 & 0.5223 & 0.6308 \\ 
& FedE &                         
\underline{\textbf{0.5478}} & \underline{\textbf{0.4378}} & \underline{\textbf{0.6831}} & \underline{\textbf{0.7481}} & \underline{\textbf{0.4183}} & \underline{\textbf{0.2789}} & \underline{\textbf{0.5910}} & \underline{\textbf{0.6844}} & \underline{\textbf{0.3626}} & \underline{\textbf{0.2226}} & \underline{\textbf{0.5289}} & \underline{\textbf{0.6337}} \\ 

\midrule

\multirow{3}{*}{DistMulit} & Single & 
0.4893 & 0.4014 & 0.5925 & 0.6519 & 0.3655 & 0.2642 & 0.4850 & 0.5618 & 0.3465 & 0.2364 & 0.4729 & 0.5703 \\ 
& Entire & 
0.5167 & 0.4186 & 0.6334 & 0.7006 & 0.3882 & 0.2772 & 0.5225 & 0.5988 & 0.3391 & 0.2287 & 0.4645 & 0.5644 \\ 
& FedE &                         
\underline{\textbf{0.5356}} & \underline{\textbf{0.4441}} & \underline{\textbf{0.6466}} & \underline{\textbf{0.7025}} & \underline{\textbf{0.4044}} & \underline{\textbf{0.2913}} & \underline{\textbf{0.5416}} & \underline{\textbf{0.6240}} & \underline{\textbf{0.3745}} & \underline{\textbf{0.2568}} & \underline{\textbf{0.5111}} & \underline{\textbf{0.6146}} \\ 

\midrule

\multirow{3}{*}{ComplEx} & Single & 
0.4843 & 0.3948 & 0.5903 & 0.6532 & 0.3723 & 0.2691 & 0.4914 & 0.5780 & 0.3262 & 0.2156 & 0.4496 & 0.5566 \\ 
& Entire & 
0.5004 & 0.4068 & 0.6164 & 0.6745 & 0.3689 & 0.2633 & 0.4902 & 0.5742 & 0.3340 & 0.2238 & 0.4541 & 0.5545 \\ 
& FedE &                         
\underline{\textbf{0.5286}} & \underline{\textbf{0.4349}} & \underline{\textbf{0.6418}} & \underline{\textbf{0.7016}} & \underline{\textbf{0.4094}} & \underline{\textbf{0.2981}} & \underline{\textbf{0.5440}} & \underline{\textbf{0.6318}} & \underline{\textbf{0.3610}} & \underline{\textbf{0.2449}} & \underline{\textbf{0.4955}} & \underline{\textbf{0.5952}} \\ 

\midrule

\multirow{3}{*}{RotatE} & Single & 
0.5271 & 0.4265 & 0.6470 & 0.7063 & 0.4062 & 0.2844 & 0.5559 & 0.6418 & 0.3660 & 0.2479 & 0.4994 & 0.6016 \\ 
& Entire & 
\textbf{0.5570} & \textbf{0.4494} & 0.6866 & 0.7510 & 0.4212 & 0.2886 & 0.5824 & 0.6733 & 0.3714 & 0.2399 & 0.5236 & 0.6349 \\ 
& FedE &                         
\underline{0.5553} & \underline{0.4436} & \underline{\textbf{0.6909}} & \underline{\textbf{0.7542}} & \underline{\textbf{0.4297}} & \underline{\textbf{0.2930}} & \underline{\textbf{0.6004}} & \underline{\textbf{0.6851}} & \underline{\textbf{0.3887}} & \underline{\textbf{0.2571}} & \underline{\textbf{0.5450}} & \underline{\textbf{0.6503}} \\  

\bottomrule
\end{tabular}}

\caption{Results on FB15k-237-Fed3. \textbf{Bold} number denotes the best result of all settings and \underline{underline} number denotes that better result between \textit{FedE} setting and \textit{Single} setting.}
\label{tab:overall-result-fb15k237-fed3}
\end{table*}

\begin{table*}[t]
\linespread{1.3}
\centering
\resizebox{\textwidth}{31mm}{
\begin{tabular}{ccllll|llll|llll}
\toprule

\multirow{2}{*}{KGE} & 
\multirow{2}{*}{Setting} & 
\multicolumn{4}{c}{Client 1} & \multicolumn{4}{c}{Client 2} & \multicolumn{4}{c}{Client 3} \\ 
\cmidrule{3-14}
& & \multicolumn{1}{c}{MRR} & \multicolumn{1}{c}{H@1} & \multicolumn{1}{c}{H@5} & \multicolumn{1}{c}{H@10} & \multicolumn{1}{c}{MRR} & \multicolumn{1}{c}{H@1} & \multicolumn{1}{c}{H@5} & \multicolumn{1}{c}{H@10} & \multicolumn{1}{c}{MRR} & \multicolumn{1}{c}{H@1} & \multicolumn{1}{c}{H@5} &  H@10 \\ 

\midrule

\multirow{3}{*}{TransE} & Single & 
0.4550 & 0.3048 & 0.6459 & 0.7390 &   0.2411 & 0.1084 & 0.3832 & 0.4895 &   0.2375 & 0.1154 & 0.3650 & 0.4673 \\ 
& Entire & 
0.4804 & 0.3327 & 0.6627 & 0.7636 &   0.2724 & 0.1349 & 0.4262 & 0.5296 &   0.2800 & \textbf{0.1560} & 0.4173 & 0.5181 \\ 
& FedE &                         
\underline{\textbf{0.4915}} & \underline{\textbf{0.3421}} & \underline{\textbf{0.6766}} & \underline{\textbf{0.7783}} &   \underline{\textbf{0.2671}} & \underline{\textbf{0.1189}} & \underline{\textbf{0.4210}} & \underline{\textbf{0.5415}} &   \underline{\textbf{0.2846}} & \underline{0.1484} & \underline{\textbf{0.4342}} & \underline{\textbf{0.5440}} \\ 

\midrule

\multirow{3}{*}{DistMulit} & Single & 
0.4084 & 0.2950 & 0.5430 & 0.6425 & 0.2129 & 0.1411 & 0.2815 & 0.3555 & 0.2326 & 0.1779 & 0.2866 & 0.3339 \\ 
& Entire & 
0.4520 & 0.3257 & 0.6069 & 0.7061 & \textbf{0.2870} & \textbf{0.2025} & 0.3727 &  0.4558 & \textbf{0.2943} & \textbf{0.2185} & \textbf{0.3690} & \textbf{0.4347} \\ 
& FedE &                         
\underline{\textbf{0.4843}} & \underline{\textbf{0.3582}} & \underline{\textbf{0.6425}} & \underline{\textbf{0.7307}} & \underline{0.2824} & \underline{0.1946} & \underline{\textbf{0.3732}} & \underline{\textbf{0.4620}} & \underline{0.2737} & \underline{0.1976} & \underline{0.3515} & \underline{0.4129} \\ 

\midrule

\multirow{3}{*}{ComplEx} & Single & 
0.4213 & 0.2967 & 0.5728 & 0.6811 &   0.2566 & 0.1702 & 0.3457 & 0.4305 &   0.2283 & 0.1625 & 0.2875 & 0.3600 \\ 
& Entire & 
0.4585 & 0.3319 & 0.6072 & 0.7169 &   0.3034 & 0.2142 & 0.3959 & 0.4833 &   0.3328 & 0.2524 & 0.4100 & 0.4848 \\ 
& FedE &                         
\underline{\textbf{0.4938}} & \underline{\textbf{0.3641}} & \underline{\textbf{0.6554}} & \underline{\textbf{0.7475}} &   \underline{\textbf{0.3336}} & \underline{\textbf{0.2378}} & \underline{\textbf{0.4367}} & \underline{\textbf{0.5339}} &   \underline{\textbf{0.3367}} & \underline{\textbf{0.2528}} & \underline{\textbf{0.4215}} & \underline{\textbf{0.5011}} \\ 

\midrule

\multirow{3}{*}{RotatE} & Single & 
0.4961 & 0.3723 & 0.6469 & 0.7363 &   0.3187 & 0.2221 & 0.4202 & 0.5057 &   0.3034 & 0.2101 & 0.3989 & 0.4857 \\ 
& Entire & 
0.5215 & 0.3902 & 0.6850 & 0.7779 & \textbf{0.3775} & \textbf{0.2732} & 0.4873 & 0.5903 &   0.3785 & \textbf{0.2766} & 0.4871 & 0.5806 \\ 
& FedE &                         
\underline{\textbf{0.5312}} & \underline{\textbf{0.3927}} & \underline{\textbf{0.7042}} & \underline{\textbf{0.7977}} & 
\underline{0.3771} & \underline{0.2634} & \underline{\textbf{0.5057}} & \underline{\textbf{0.5984}} & \underline{\textbf{0.3828}} & \underline{0.2733} & \underline{\textbf{0.4984}} & \underline{\textbf{0.5992}} \\ 

\bottomrule
\end{tabular}}

\caption{Results on NELL-995-Fed3.}
\label{tab:overall-result-nell995-fed3}
\end{table*}

The results on FB15k-237-Fed3 and NELL-995-Fed3 are summarized in Table \ref{tab:overall-result-fb15k237-fed3} and Table \ref{tab:overall-result-nell995-fed3}. 

Based on these results, we can find that our proposed FedE can achieve better link prediction performance than knowledge graph embeddings trained in \textit{Single} setting, specifically, using RotatE as KGE method, FedE is relatively improved by an average of 5.7\%, 3.6\%, 7.9\% and 7.1\% on MRR, Hits@1, Hits@5, Hits@10 among clients in FB15k-237-Fed3 and 14.2\%, 14.0\%, 15.4\% and 14.4\% among clients in NELL-995-Fed3.
Note that unless otherwise specified, in the following part, the average results on metrics among clients in one dataset means the average weighted by the number of evaluation triples in each client of this dataset.

We also find that our proposed FedE can achieve better performance than knowledge graph embedding trained in \textit{Entire} setting in most cases, and get comparable results in other cases.

Overall, For question 1) and 2), FedE can achieve better performance than embeddings trained in \textit{Single} setting stably, and better results than embeddings trained in \textit{Entire} setting, which even violates the data privacy, in most cases, indicating that the effectiveness of our FedE in learning knowledge graph embeddings in federated setting.

\begin{figure}[t]
\centering
\includegraphics[scale=0.55]{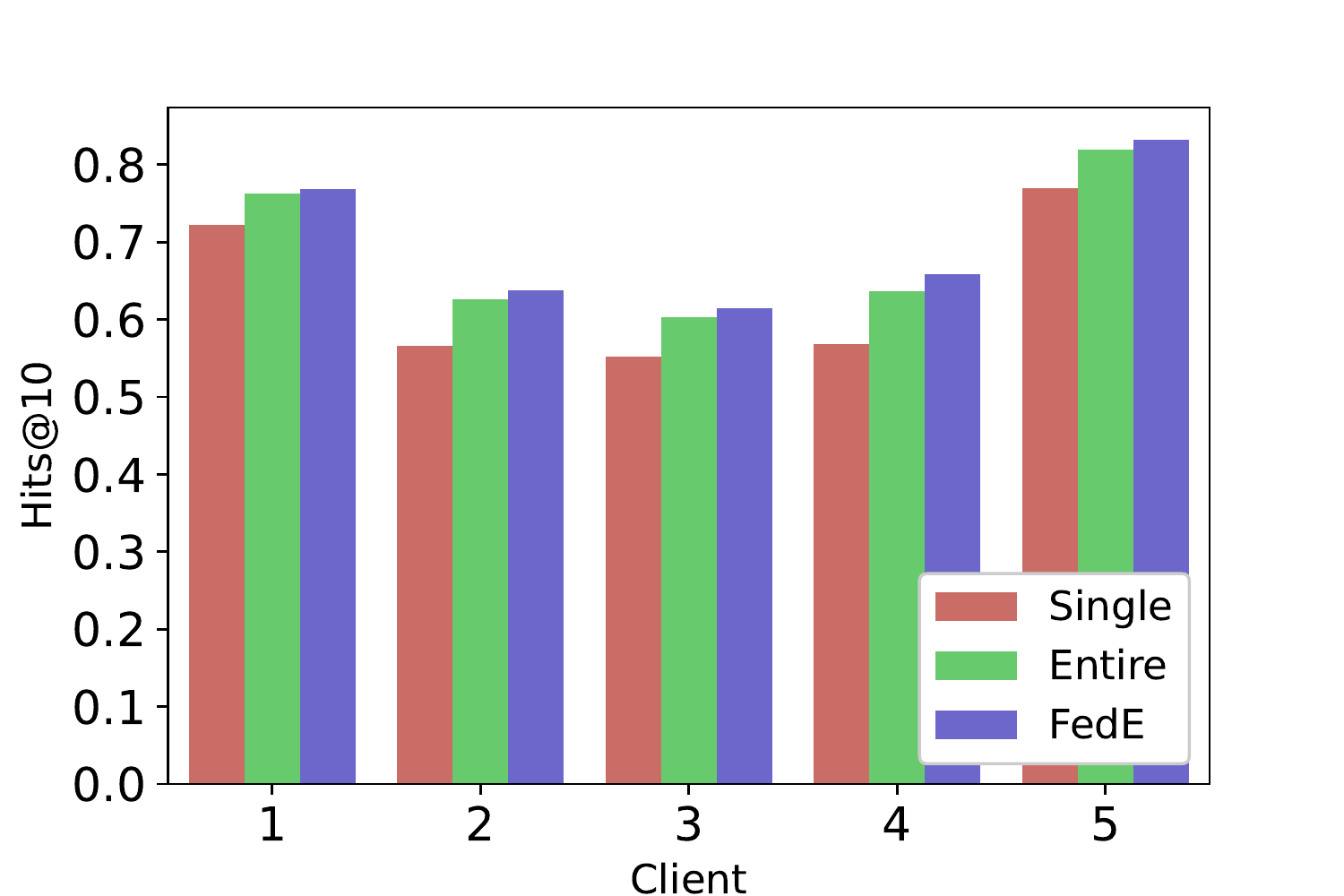}
\caption{Results of Hits@10 on FB15k-237-Fed5 using RotatE as KGE model.}
\label{fig:fed5-rst}
\end{figure}

\begin{figure}[t]
\centering
\includegraphics[scale=0.55]{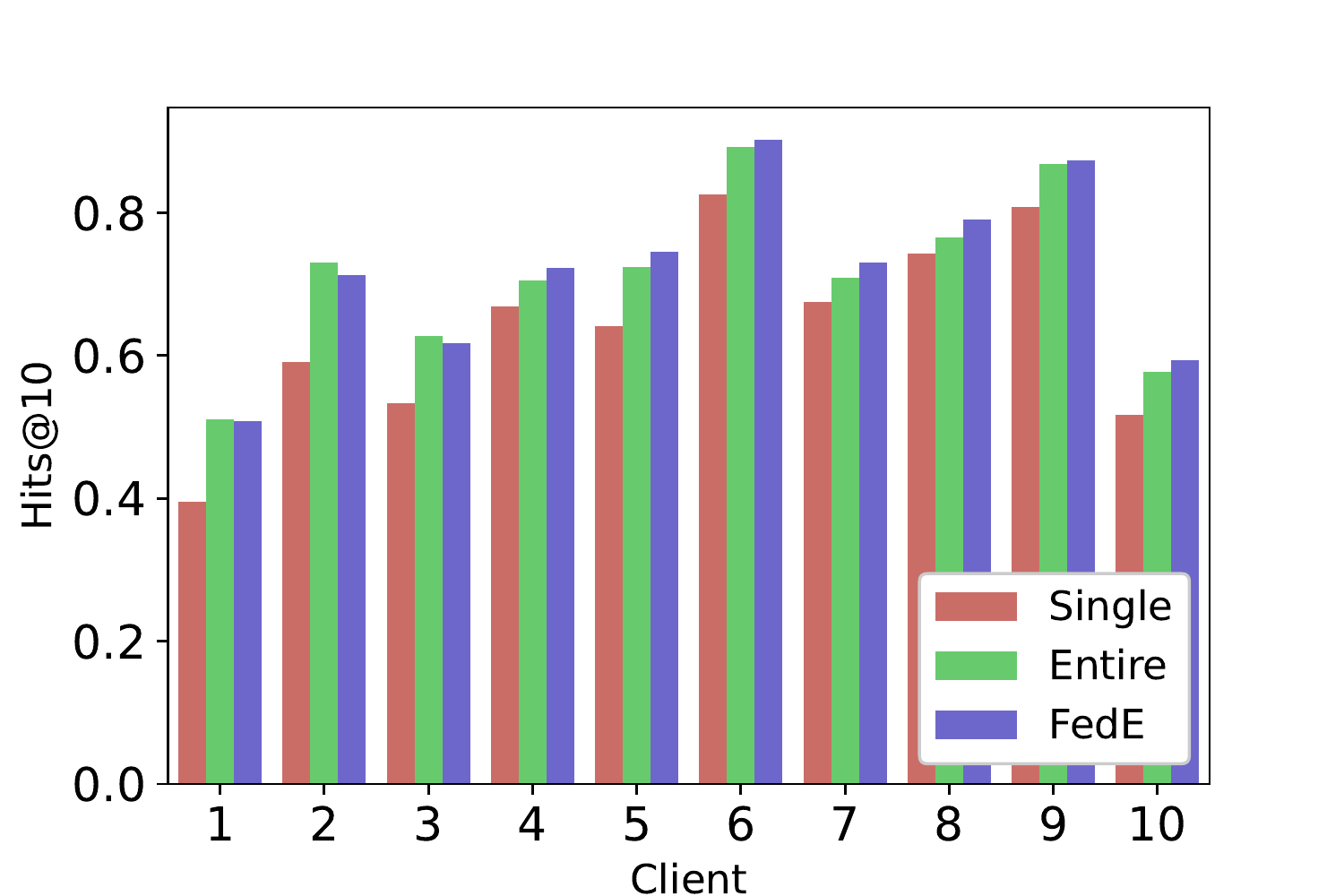}
\caption{Results of Hits@10 on FB15k-237-Fed10 using RotatE as KGE model.}
\label{fig:fed10-rst}
\end{figure}

\textbf{The Number of Clients}. In this part, we want to figure out that with the increase of the number of clients, is FedE still effective? The Hits@10 results on FB15k-237-Fed5 and FB15k-237-Fed10 using RotatE as KGE model are described in Figure \ref{fig:fed5-rst} and Figure \ref{fig:fed10-rst}. 

FedE is relatively improved by an average of 6.7\%, 1.7\%, 10.7\% and 10.3\% on MRR, Hits@1, Hits@5, Hits@10 among clients in FB15k-237-Fed5 and 7.8\%, 1.7\%, 13.1\% and 12.2\% among clients in FB15k-237-Fed10 compared with RotatE in \textit{Single} setting. For question 3), our proposed FedE is still effective with the increase of the number of clients.

\textbf{Ablation Study on Model Fusion.} As described in Section \ref{sec:model-fusion}, the procedure of model fusion in FedE can fusion the model learned with and without federated setting, i.e. take advantage of both embeddings trained under \textit{FedE} setting and \textit{Single} setting. Here we use a ablation study to show the effectiveness of model fusion procedure. We show the average MRR results among clients of these four datasets respectively in Table \ref{tab:ablation}. For question 4), we find that the link prediction results have a steady improvement on different datasets with model fusion, which shows the effectiveness of model fusion.

\begin{table}[t]
\linespread{1.4}
\small
\centering

\begin{tabular}{ccc}
\toprule

Dataset & w/o Fusion & w/ Fusion \\

\midrule

FB15k-237-Fed3 & 0.4434 & 0.4543 \\
FB15k-237-Fed5 & 0.4345 & 0.4439 \\
FB15k-237-Fed10 & 0.4208 & 0.4340 \\
NELL-995-Fed3 & 0.4289 & 0.4433 \\

\bottomrule
\end{tabular}

\caption{Ablation study on model fusion.}
\label{tab:ablation}
\end{table}

\subsection{Factors That Have Impact on FedE}

\subsubsection{Multi-Client Parallelism}

In this part, we focus on exploring the impact of multi-client parallelism (i.e. the value of client fraction $F$ in each round) on learning of FedE. We train several FedE models on FB15k-237-Fed10 by varying client fraction $F$ as \{0.1, 0.3, 0.5, 0.8, 1\} using RotatE as KGE model, and check the performance on validation set each 5 round.

\begin{figure}[t]
\centering
\includegraphics[scale=0.54]{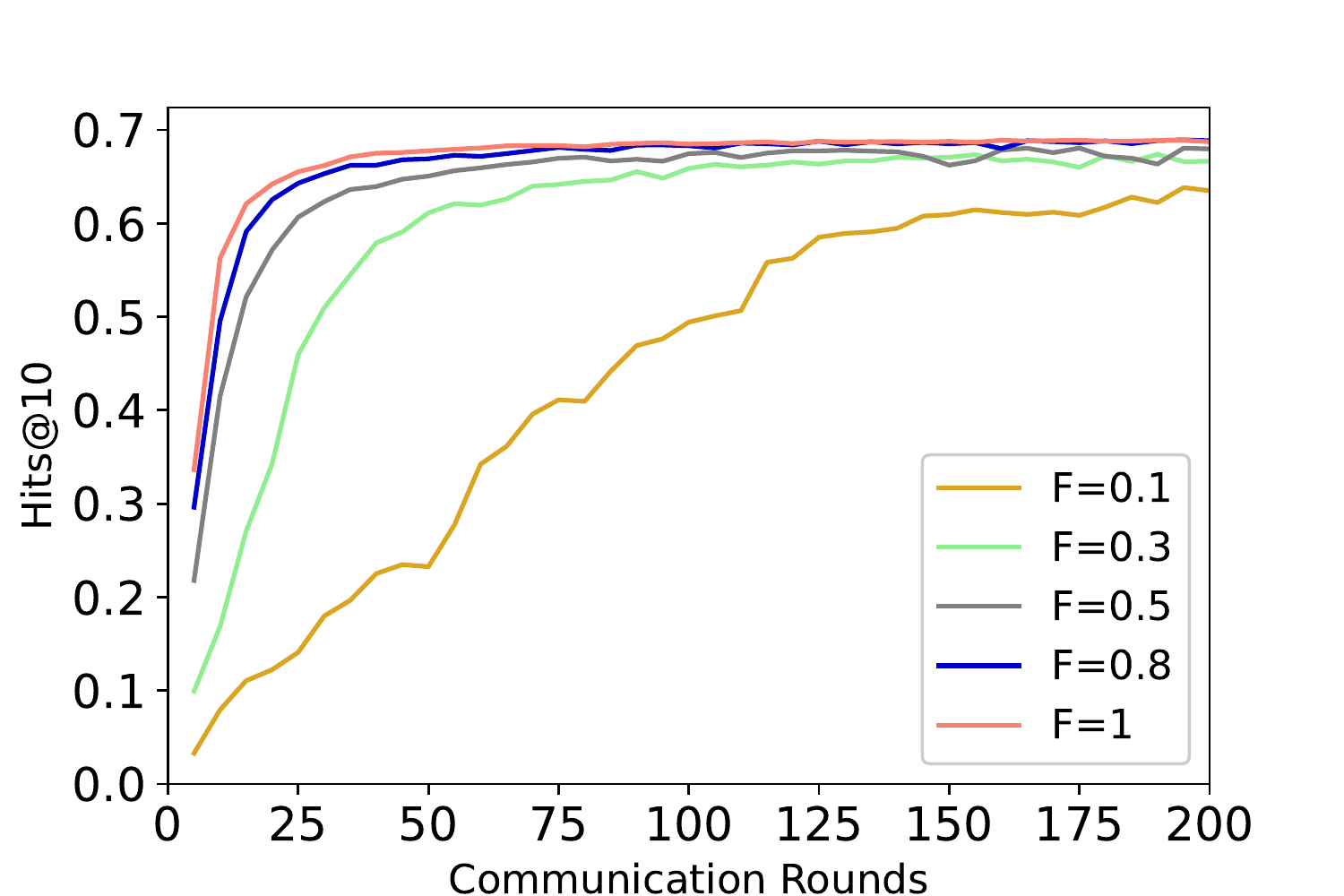}
\caption{Average Hits@10 on validation set over clients of FB15k-237-Fed10 versus communication rounds.}
\label{fig:client-paral}
\end{figure}

We report the change of average Hits@10 on validation set over clients of FB15k-237-Fed10 versus communication rounds in Figure \ref{fig:client-paral}, and different curves means using different client fractions for FedE. We find that increasing the multi-client parallelism can boost the convergence rate.

\begin{figure}[t]
\centering
\includegraphics[scale=0.33]{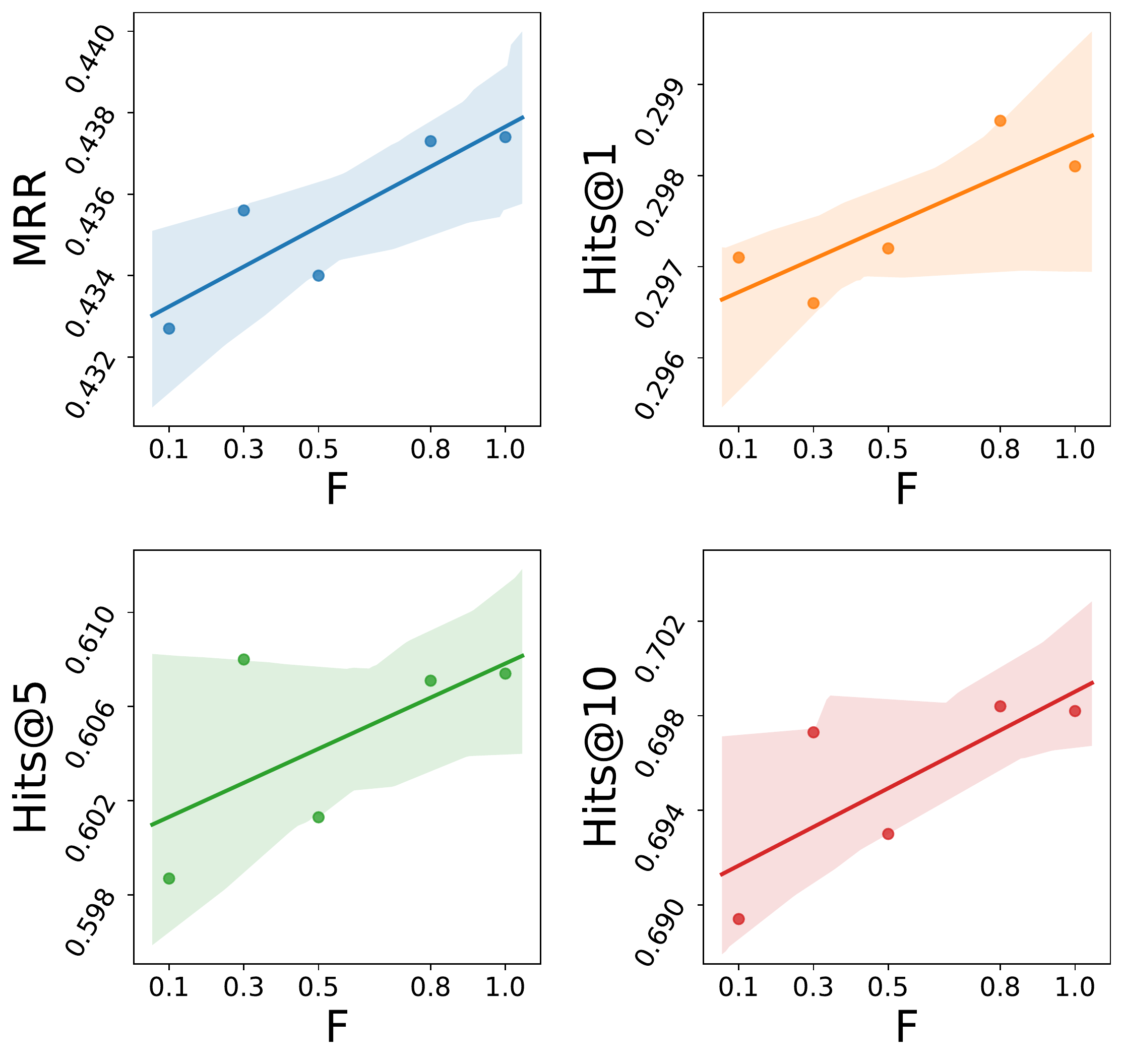}
\caption{Average test results of FB15k-237-Fed10 with different $F$.}
\label{fig:client-paral-test}
\end{figure}

Furthermore, we also evaluate trained FedE with different client fractions on test set of FB15k-237-Fed10, and show the average test results of different metrics in Figure \ref{fig:client-paral-test}. Each sub-figure in Figure \ref{fig:client-paral-test} shows a specific metric of average test result vs. client fraction, and the lines are fitted by linear regression model. The translucent bands around the regression lines demonstrate the confidence interval for regression estimate. From theses results, we find that increasing the multi-client parallelism during training of FedE can improve the performance of FedE.

\subsubsection{Client Computation}

\begin{figure}[t]
\centering
\includegraphics[scale=0.54]{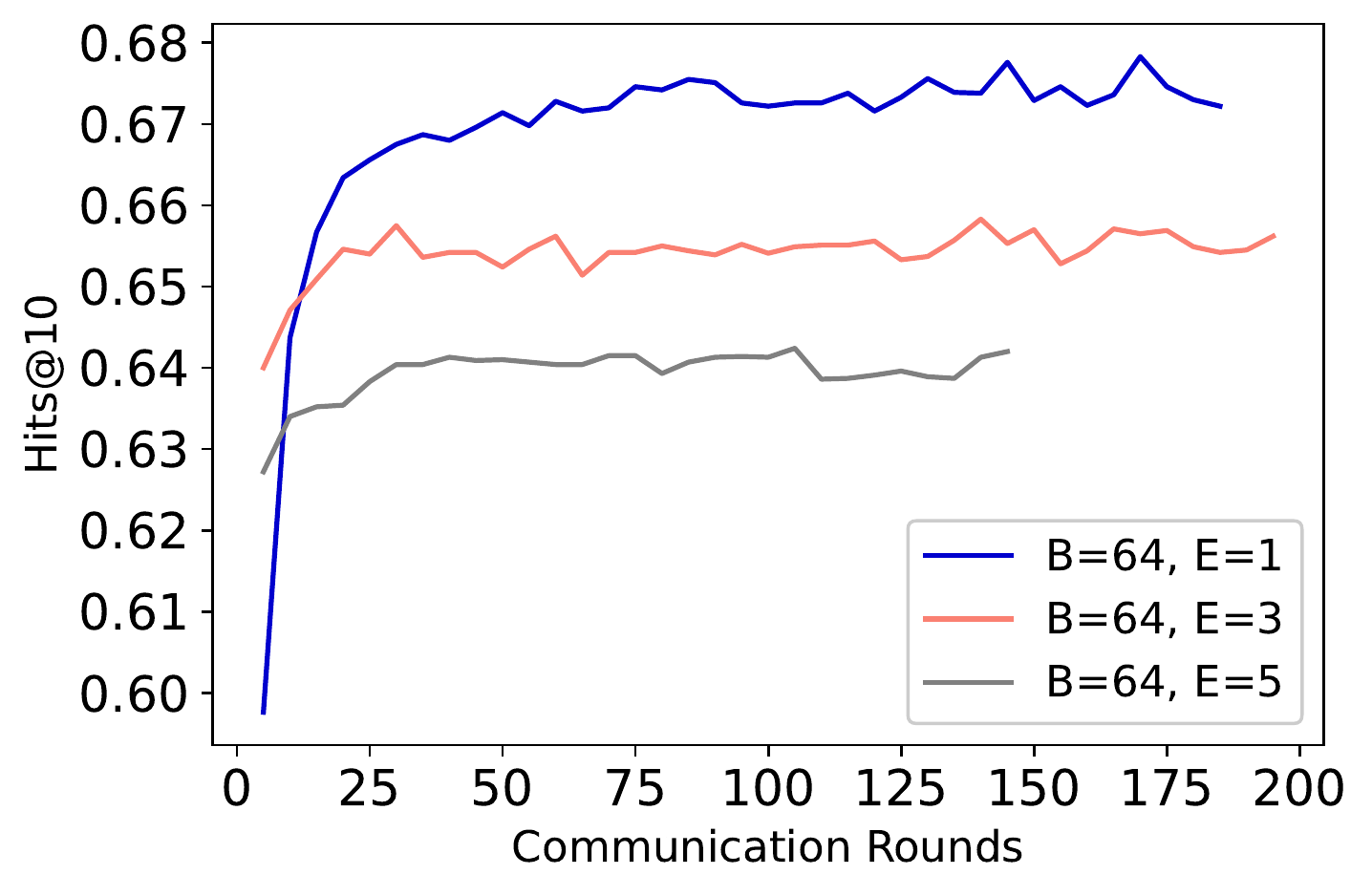}
\caption{The average validation Hits@10 result curves with fixing local batch size and varying local epoch on FB15k-237-Fed10.}
\label{fig:comp-epoch-bs64}
\end{figure}

\begin{figure}[t]
\centering
\includegraphics[scale=0.54]{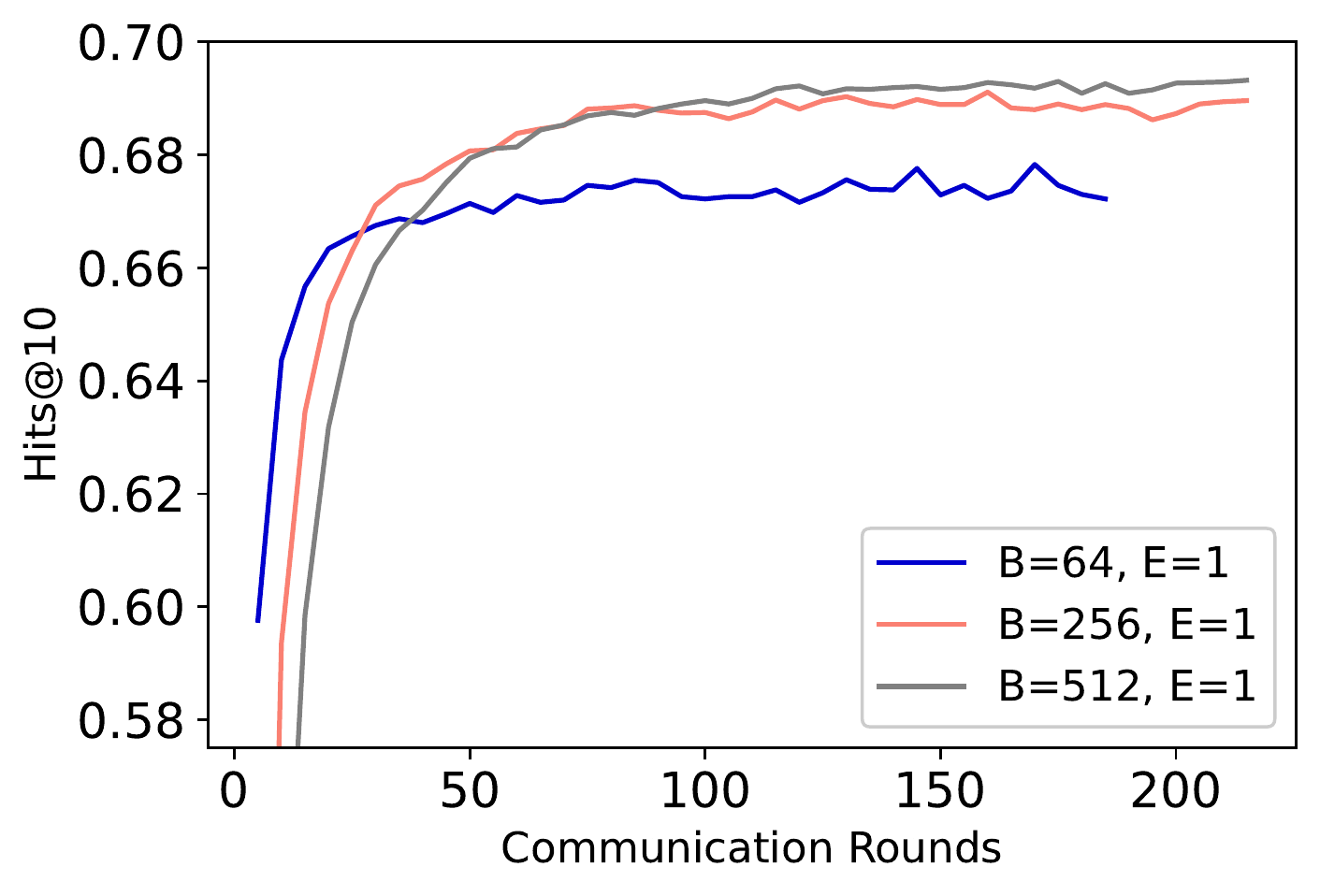}
\caption{The average validation Hits@10 result curves with fixing local epoch and varying local batch size on FB15k-237-Fed10.}
\label{fig:comp-bs-epoch1}
\end{figure}

Client computation (i.e. the number of local updating) is also an important factor should be considered in learning knowledge graph embeddings in federated setting, because clients in such setting sometimes may have weak computing power like mobile device. Client computation is control by local epoch $E$ and local batch size $B$, and the value of which is proportional to $E/B$, in other words, more local epoch and smaller batch size make more client computation. In this section, to investigate the impact of client computation on convergence rate and model performance, we train FedE with RotatE as KGE model on FB15k-237-Fed10, and fix the client fraction $F$ as 0.5, varying local batch size $B$ with \{64, 256 ,512\} and local epoch $E$ with \{1, 3, 5\}. 

The curves of average validation Hits@10 results during training with fixing local batch size and varying local epoch are described in Figure \ref{fig:comp-epoch-bs64}. From such curves, we can find that smaller local epoch value can make better model performance while reduce the convergence rate of FedE learning.

The curves of average validation Hits@10 results during training with fixing local epoch and varying local batch size are described in Figure \ref{fig:comp-bs-epoch1}. From such curves, we can find that larger batch size can make better model performance while reduce the convergence rate.

Finally, based on what we observer above, we can find that when we try to reduce the number of local updating (client computation), i.e. reduce local epoch and/or increase local batch size, we can get better performance while may have a lower convergence rate which makes more communication costs between server and clients.

\section{Related Work}

\subsection{Knowledge Graph Embedding}

To overcome the disadvantage (e.g. low efficiency and robustness) of using symbolic methods to represent knowledge graphs, researchers seek to represent relations and entities as continuous vectors, so as to mapping the operation of knowledge graphs into vector spaces. Such vector representation methods (i.e. Knowledge Graph Embedding, KGE) mainly focus on defining score functions to expressively measure the truth value of triples in knowledge graphs. A lot of KGE methods \cite{kgembedding} are proposed in the last decade and applied to various applications \cite{recsys, qa, MetaR, WRAN}, most of them can be classified as translation-based method, tensor factorization method and neural-network-based method.

TransE \cite{TransE} is the first translation-based method which defines score function based on interpreting relations as translation operations on embeddings of the entities. Following TransE, TransH \cite{TransH} introduces relation-specific hyperplanes to overcome the flaws of TransE, e.g. dealing with 1-N, N-1 and N-N relations. TransR \cite{TransR} and TransD \cite{TransD} are extension methods sharing the general idea with above methods. RotatE \cite{RotatE} maps entities and relations into complex vector space and treat relations as rotations between entities, which improves the ability of pattern modeling.

RESCAL \cite{RESCAL} learns embeddings by performing collective tensor factorization of a three-way tensor. Following RESCAL, DistMult \cite{DistMult} restricts relation matrices as diagonal matrices. Further, ComplEx \cite{ComplEx} is an extension of DistMult using complex valued embeddings for modeling antisymmetric relations.

Neural-network-based methods including methods based on architecture like neural network, convolutional neural network and graph neural network. NTN \cite{NTN} proposes a neural tensor network to define the operation between head, tail entity and relation for scoring triples. ConvE \cite{ConvE} uses 2D convolutions over embeddings to define the truth value of triples. As the graph nature of knowledge graphs, it is intuitive to leverage Graph Neural Network which has proven to be effective for graph structure to embed entities and relations into vector spaces, for example, R-GCN \cite{RGCN} and CompGCN \cite{CompGCN} proposed KGE methods based on Graph Convolutional Networks, and KBGAT \cite{KBGAT} and A2N \cite{A2N} based on Graph Attention Networks.

\subsection{Federated Learning}

Recently, \citet{FedAvg} proposed Federated Learning, trying to learn a shared model by data in several clients/platforms while not aggregating data together for privacy protection. Many researchers in different fields have proposed methods in federated setting. \citet{FedMeta} proposed a federated meta-learning framework by sharing parameterized algorithm instead global model. \citet{FedNER} proposed FedNER to handle medical NER task under federated setting. \citet{fedre} explores distant supervised relation extraction in federated setting. \citet{FTM} proposed federated topic modeling framework for collaboratively learning a shared topic model while maintaining privacy protection from data from multiple parties.

\section{Conclusion}

We propose a federated knowledge graph embedding framework FedE, in federated setting. FedE learns knowledge graph embeddings for each client by taking advantage of triples from other clients while not collecting triples together for data privacy. The proposed FedE has impressive improvement compared with trained embedding models based only on its own knowledge graph for each client, i.e. the \textit{Single} setting. Furthermore, we also analyze the impact of multi-client parallelism and client computation on training and testing of FedE.


\bibliographystyle{acl_natbib}
\bibliography{emnlp2020}

\appendix


\end{document}